\documentclass[10pt,twocolumn,letterpaper]{article}

\usepackage{cvpr}
\usepackage{times}
\usepackage{epsfig}
\usepackage{graphicx}
\usepackage{amsmath}
\usepackage{amssymb}
\usepackage{booktabs}
\usepackage{comment}
\usepackage{dsfont}
\usepackage{microtype}
\usepackage{float}
\usepackage[ruled,linesnumbered,lined,commentsnumbered]{algorithm2e}
\usepackage[subrefformat=parens]{subcaption}

\usepackage[hyphens]{url}
\usepackage[pagebackref=true,breaklinks=true,colorlinks,bookmarks=false]{hyperref}

\makeatletter
\newcommand{\printfnsymbol}[1]{%
  \textsuperscript{\@fnsymbol{#1}}%
}

\makeatother

\cvprfinalcopy % *** Uncomment this line for the final submission
\def\cvprPaperID{8114} % *** Enter the CVPR Paper ID here

\begin{document}
\title{3rd Place Solution to Meta AI Video Similarity Challenge}

\author{Shuhei Yokoo\\
LINE Corporation\\
{\tt\small shuhei.yokoo@linecorp.com}
\and
Peifei Zhu\\
LINE Corporation\\
{\tt\small peifei.zhu@linecorp.com}
\and
Junki Ishikawa\\
LINE Corporation\\
{\tt\small junki.ishikawa@linecorp.com}
\and
Rintaro Hasegawa\\
Keio University\\
{\tt\small hasegawa@ailab.ics.keio.ac.jp}
}

% \author{
% Shuhei Yokoo, Peifei Zhu, Junki Ishikawa, Rintaro Hasegawa\\
% LINE Corporation\\
% {\tt\small \{shuhei.yokoo, peifei.zhu, junki.ishikawa, rintaro.hasegawa\}@linecorp.com}
% }

\maketitle

\begin{abstract}
This paper presents our 3rd place solution in both Descriptor Track and Matching Track of the Meta AI Video Similarity Challenge (VSC2022), a competition aimed at detecting video copies. Our approach builds upon existing image copy detection techniques and incorporates several strategies to exploit on the properties of video data, resulting in a simple yet powerful solution. By employing our proposed method, we achieved substantial improvements in accuracy compared to the baseline results (Descriptor Track: 38\% improvement, Matching Track: 60\% improvement).
 Our code is publicly available here: \url{https://github.com/line/Meta-AI-Video-Similarity-Challenge-3rd-Place-Solution}
\end{abstract}

\begin{figure}[t]
\centering
% \vspace{0.3mm}
  \includegraphics[width=0.95\linewidth]{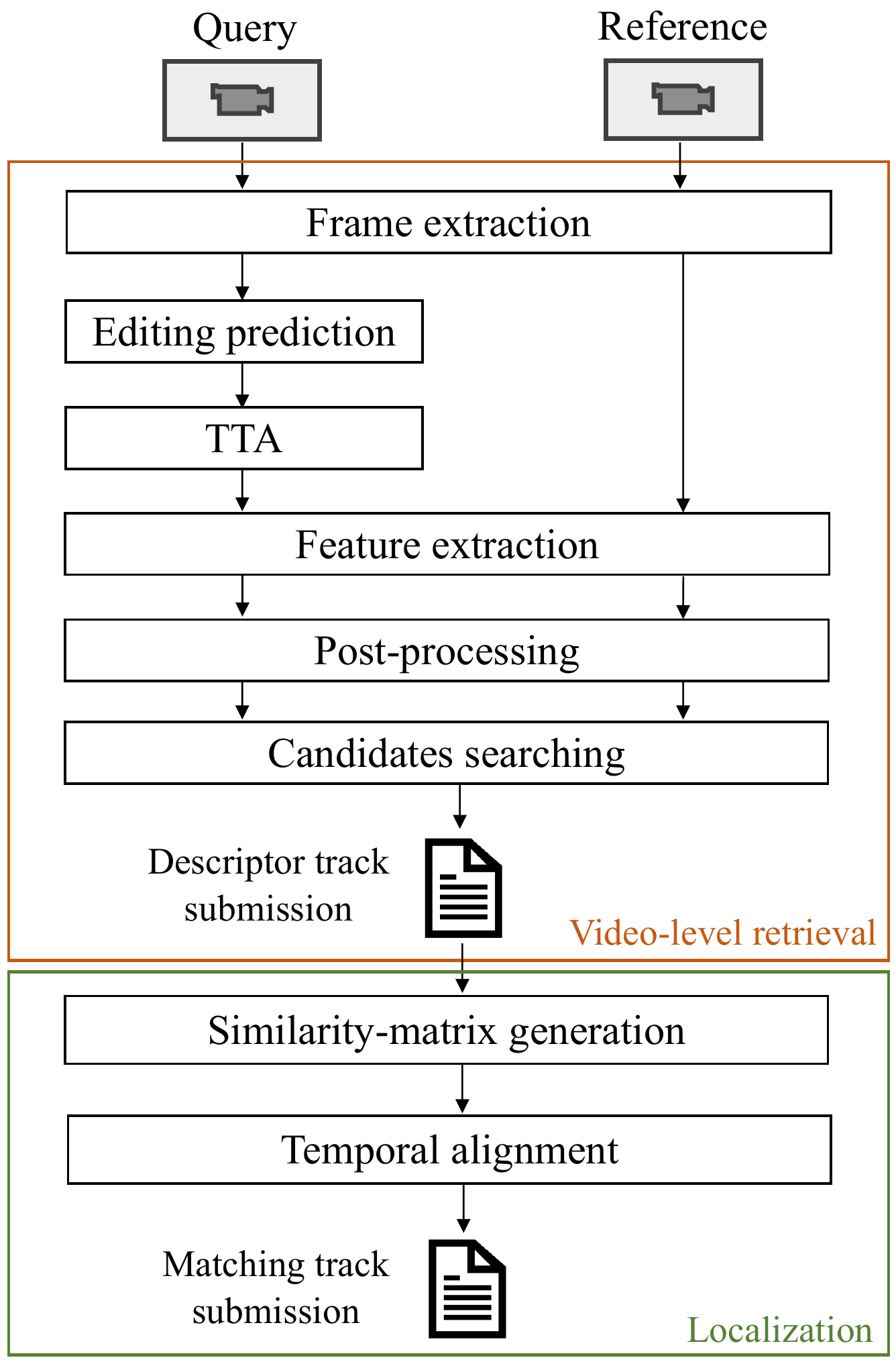}
  % \vspace{-0.3mm}
  \caption{
    Our solution pipeline overview of video copy detection for inference. The feature extraction is mainly based on an image copy detection model (ISC-dt1~\cite{ISC-dt1}), and the temporal alignment is based on Temporal Network~\cite{Tan2009ScalableTN}. We implement several tricks to improve the detection performance.
  }
  \label{fig:framework}
\end{figure}

\begin{figure*}[t]
\centering
% \vspace{0.3mm}
  \includegraphics[width=0.8\linewidth]{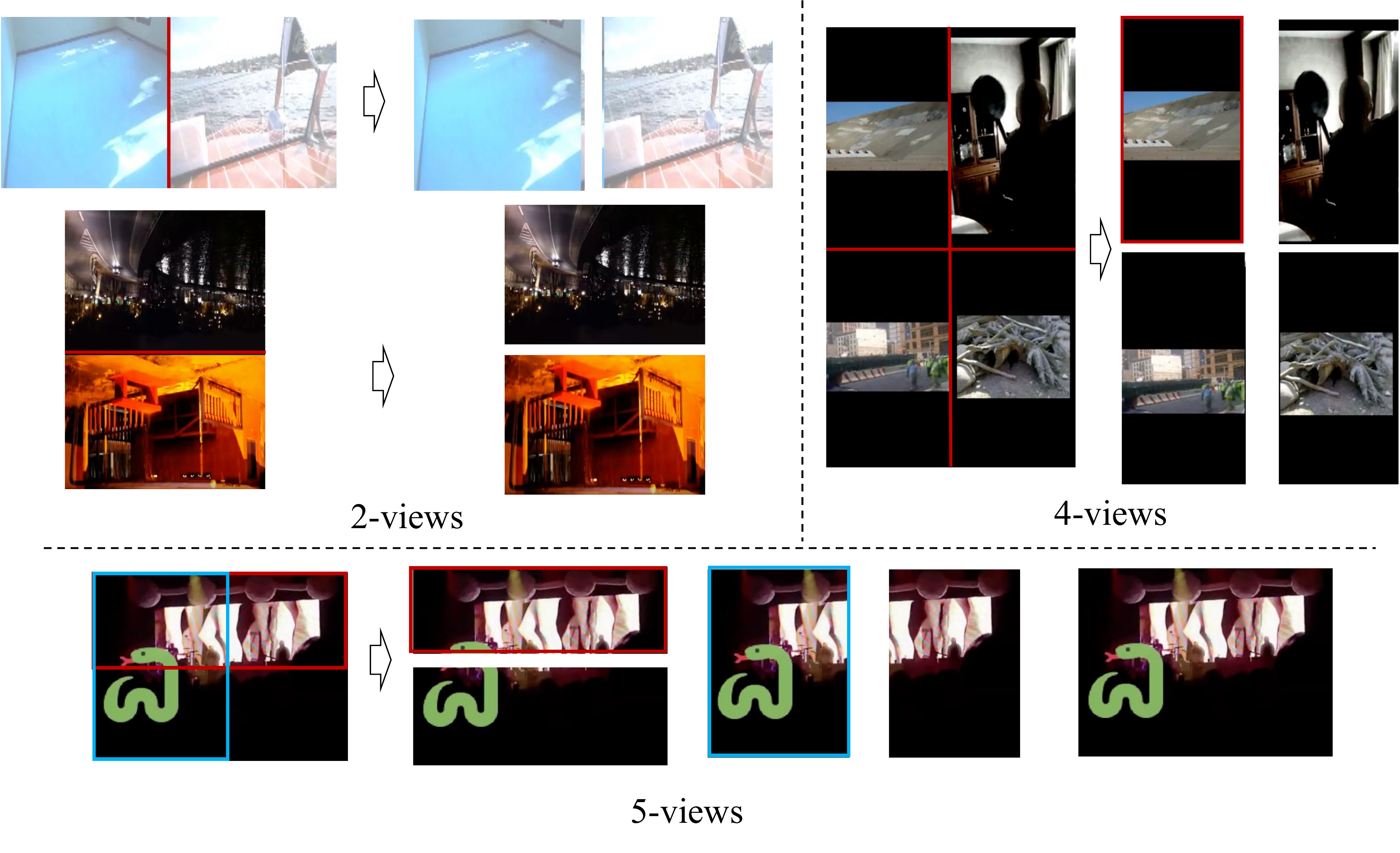}
%   \vspace{0.3mm}
  \caption{
    The three types of multi-crop used in our method. 2-views crop is for vertical or horizontal stacking, 4-views crop is for vertical plus horizontal stacking, and 5-views is for all other manipulations. Multi-crop enables the query videos to be matched both globally and locally and significantly improves the detection results. Credit of the original videos in this figure: (user name
of Flickr): ``madc0w'', ``bamblue88'', ``SupremeCrete'', ``Wam Kat'', ``permanently scatterbrained'', ``jfgornet'', ``Salim Virji'', ``scott.fuhrman'', ``terryballard'', ``FiendishX'', ``Steven Smith!'', ``tristanf'', ``itsbeach''.
  }
  \label{fig:multi-crop}
\end{figure*}

\section{Introduction}
In recent years, with the rapid development of social media, issues such as plagiarism and unauthorized modification have become increasingly severe. Consequently, there is a growing demand for technologies capable of accurately and automatically detecting illegal content. In response to this situation, the field of copy detection has experienced a swift expansion in recent years, with research on image and video copy detection~\cite{SSCD2022,wang2023benchmark,fernandez2022active,Jiang2014VCDBALCopy,he2022largeVCSL,he2023transvcl} gaining significant attention. Notably, the Image Similarity Challenge at NeurIPS'21 (ISC2021) was organized as a competition for image copy detection~\cite{ISC2021,papakipos2022resultsISC}, followed by the recent organization of the Meta AI Video Similarity Challenge at CVPR’23 (VSC2022) focusing on video copy detection.

VSC2022 has two tracks: descriptor and matching. In the descriptor track, the task is to compute embeddings (up to 512 dimensions) for each video. In the matching track, the task is to generate predicted matches containing the starting and ending timestamps of query-reference video pairs.

In this paper, we describe our approach that achieved third place in both tracks of the VSC2022 competition. Our solution can be summarized as follows:

\begin{enumerate}
    \item We propose a simple yet powerful video copy detection pipeline, based on an image copy detection model (ISC-dt1~\cite{ISC-dt1}).
    \item Our approach utilizes test time augmentation based on the predictions of an editing prediction model.
    \item We exploit the properties of videos through various techniques: emphasizing copy videos more in search results by calculating frame consistency, concatenating adjacent frames followed by dimensionality reduction using PCA, and localizing copied frames by employing Temporal Network~\cite{Tan2009ScalableTN}.
\end{enumerate}

\section{Method}
\subsection{Pipeline Overview} 
As previous works have proved frame-level features are necessary to precisely locate copied segments and achieved better performance in video copy detection tasks~\cite{kordopatis2019visil, shao2021temporal}, our method also extracts frame-level features to retrieve copied pairs and applies video temporal alignment to localize copied segments. The overview of our method is shown in Figure~\ref{fig:framework}. Details will be described in the following subsections.

In video-level retrieval, the embeddings of query and reference videos are extracted respectively. For query videos, since they can either be original videos (without any copied segments) or edited videos (with copied segments), we first implement a process to predict whether they are edited. Only videos being predicted as edited will be processed in the following steps. Next, multiple crop which is a type of test time augmentation (TTA), is performed so that the query frames can be matched both globally and locally. On the other hand, since reference videos do not contain any copied segments or manipulations, they are directly passed to the feature extraction once the frames are extracted. For feature extraction, we use ISC-dt1~\cite{ISC-dt1} which is the 1st place model of ISC2021 Descriptor Track. 
Post-processing including our proposed Consistency Weighting, Temporal Concat, and score normalization~\cite{SSCD2022} are applied to further improve the accuracy. The last step in video-level retrieval is to search copied pairs from the query and reference embeddings. We use an exhaustive search to obtain the top 1200 candidates for each query.

To localize the starting and ending timestamps of the copied segments between the candidate copied pairs, we generate frame-to-frame similarity matrix for each pair and apply temporal alignment. Various methods have been proposed for temporal alignment~\cite{Tan2009ScalableTN, douze2010image, chou2015pattern}, and we choose a graph-based method named Temporal Network (TN)~\cite{Tan2009ScalableTN}. TN constructs a graph network that takes matched frames as nodes and similarities between frames as weights of links, and thus the path with maximized weight indicates the location of the copied segment. Finally, the video copy pairs and their matching results are obtained. 

\subsection{Editing Prediction} 
Editing prediction aims to classify what type of manipulations are used for copy videos.
Knowing the types of manipulations helps us to choose proper augmentation methods that return the edited video to its original appearance.

We build an editing prediction model by training with simulation data.
We first create a simulated copy video dataset using the training reference of the VSC2022 data. We randomly select two videos from the training reference and copy a random segment from one video to another. To better reproduce the character of the challenge data, we use a data augmentation library named AugLy~\cite{bitton2021augly} to add manipulations including blending, altering brightness or contrast, blurring, stacking, etc. We considered editing prediction as a multi-label classification task and build a model based on ConvNeXt~\cite{liu2022convnet}. 

\subsection{Multiple Crops}
The challenge data contains manipulations that concatenate multiple videos spatially, such as stacking, and overlay. Only part of the query image can be matched to the reference image in such cases, and thus using multiple crops to match both globally and locally can be effective. We design three types of multiple crops, 2-views for vertical or horizontal stacking, 4-views for vertical plus horizontal stacking and 5-views for the other manipulations. Examples are shown in Figure~\ref{fig:multi-crop}.

The model inferences are performed on multiple crops, and multiple descriptors are extracted depending on the number of views. These are stored as separate descriptors in the temporal direction with repeated timestamps.
% , as we want to treat each of these separated views independently, considering that unrelated videos might be combined

\begin{figure}[t]
\centering
  \includegraphics[width=0.9\linewidth]{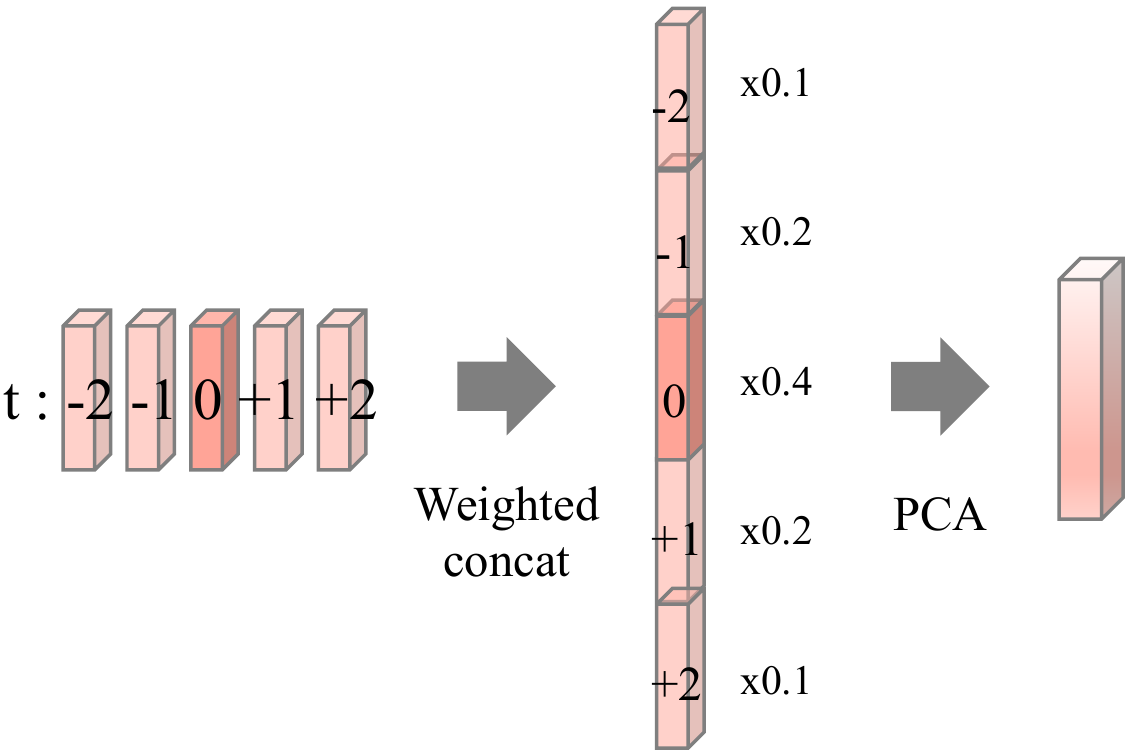}
  \vspace{1mm}
  \caption{
Overview of Temporal Concat. When concatenating descriptors, we assign larger weights to descriptors that are temporally closer to the center frame. This process is applied to all frame descriptors in a sliding-window manner.
  }
  \label{fig:temp-concat}
\end{figure}

\subsection{Post-processing}

\noindent{\bf{Consistency Weighting.}}
We analyzed the search results in the train set and noticed that there are some incorrect prediction pairs with relatively high confidence scores.
The commonality between them is that they are very similar in content, but not copies.
It is important to distinguish between videos that are copies and videos that are simply quite similar because these incorrect predictions with high scores are predominantly detrimental to the competition metric.

Therefore, we utilize the characteristics of copied videos and apply weighting to increase the confidence of copied videos. We call this method Consistency Weighting.
Specifically, in the case of copied videos, a clip from another video is inserted in the middle, causing a scene change and resulting in lower consistency between frames. Therefore, for each video descriptor, we apply weighting as represented by the following formula:

\begin{equation}
X' = \frac{X}{\frac{1}{n^2}\sum_{i=1}^n \sum_{j=1}^n (X \cdot X^T)_{ij}}, X \in \mathbb{R}^{n \times d}
\label{eq1}
\end{equation}
\vspace{0.5mm}

$X$ is the matrix representing a video, containing $d$-dimensional descriptors for $n$ frames.
This weighting scheme can increase the inner product between a copied video descriptor and reference descriptors, emphasizing the predictions of copied videos more than non-copied videos.

\vspace{2mm}
\noindent{\bf{Temporal Concat.}}
Since the frame-level features are extracted without considering any temporal information, we add a temporal post-processing to improve temporal robustness. We concatenate adjacent frame descriptors in a sliding-window manner and apply PCA to reduce their dimensions.
In this report, we refer to this temporal post-processing as Temporal Concat.
Figure~\ref{fig:temp-concat} shows an overview of the Temporal Concat method.

\begin{figure*}[t]
\centering
  \includegraphics[width=0.9\linewidth]{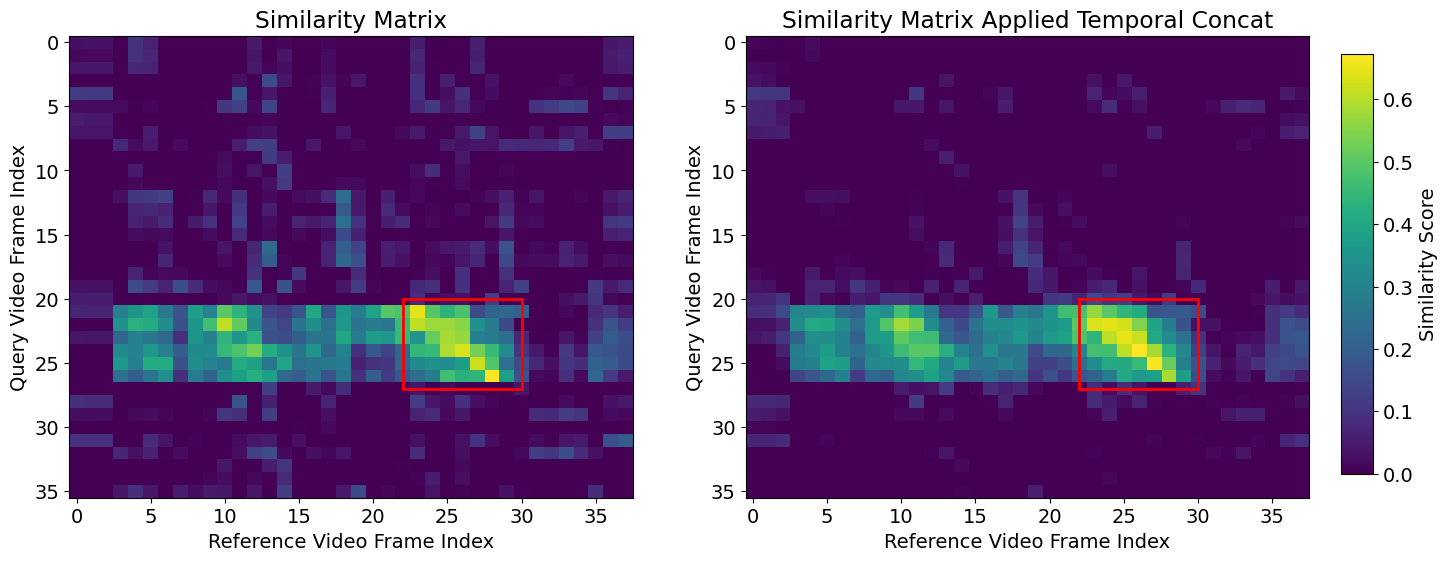}
  \caption{
Visualization of the impact of Temporal Concat on the similarity matrix. The left matrix represents the similarity matrix before applying Temporal Concat, while the right matrix represents the matrix after applying Temporal Concat. The x-axis of the visualization graph corresponds to the sequential frame numbers of the reference video, and the y-axis corresponds to the sequential frame numbers of the query video.
  }
  \label{fig:vis-sim-mat}
  \vspace{4mm}
\end{figure*}

Figure~\ref{fig:vis-sim-mat} shows the visualization of the impact of Temporal Concat on the similarity matrix for a positive pair example. The area enclosed by the red box represents the copied segment. An ideal similarity matrix would have peak values along the diagonal within the segment. After applying Temporal Concat, the similarity peaks along the diagonal become more pronounced within the copied segment, while the overall similarity outside the copied segment decreases, indicating less noise.

\begin{table*}[t]
\centering
\begin{tabular}{ccccc}
\toprule
Multi-crop & Consistency Weighting & Temporal Concat & \textmu AP (Descriptor) & \textmu AP (Matching) \\
\midrule
 &  &  & 0.6817 & 0.5706 \\
\checkmark &  &  & 0.7715 & 0.7367 \\
\checkmark & \checkmark &  & 	0.8463 & 0.7525 \\
\checkmark & \checkmark & \checkmark & 0.8692 & 0.7594 \\
\bottomrule
\end{tabular}
\vspace{0.2mm}
\caption{Ablation from baseline to our final solution for both tracks. \textmu AP (Descriptor) is a score of Descriptor Track, and \textmu AP (Matching) is a score of Matching Track.}
\label{tab:ablation}
\vspace{1mm}
\end{table*}

\begin{table}[t]
\centering
\begin{tabular}{l c c}
\toprule
Team & \textmu AP \\
\midrule
do something & 0.8717 \\
FriendshipFirst1             & 0.8514    \\
\textbf{cvl-descriptor (Ours)}         & \textbf{0.8362}   \\
\midrule
Baseline & 0.6047		\\
\bottomrule
\end{tabular}
\vspace{0.2mm}
\caption{Leaderboard with Descriptor Track final results (only top 3 teams are listed here).
Our results are in bold.}
\label{tab:lb_descriptor}
\vspace{1mm}
\end{table}

\begin{table}[t]
\centering
\begin{tabular}{l c c}
\toprule
Team & \textmu AP \\
\midrule
do something more             & 0.9153    \\
CompetitionSecond2 & 0.7711 \\
\textbf{cvl-matching (Ours)}         & \textbf{0.7036}   \\
\midrule
Baseline & 0.4411		\\
\bottomrule
\end{tabular}
\vspace{0.2mm}
\caption{Leaderboard with Matching Track final results (only top 3 teams are listed here).
Our results are in bold.}
\label{tab:lb_matching}
\vspace{1mm}
\end{table}

\vspace{10mm}
\noindent{\bf{\bf Score normalization.}}
Score normalization~\cite{ISC2021,SSCD2022} is a widely-used trick for ranking systems. It aims to make the similarity score comparable across different queries. We use the same score normalization as described in~\cite{SSCD2022}. The training reference of the VSC2022 data is used as the ``noise'' dataset.
% \vspace{2mm}

\section{Experiments}
\subsection{Dataset}
VSC2022 provides a new video copy detection dataset composed of approximately 100,000 videos derived from the YFCC100M~\cite{2016YFCC100M} dataset. The training dataset contains 8,404 query videos, 40,311 reference videos, and the ground truth for the query videos which contain content derived from reference videos. Edited query videos may have been modified using a number of techniques including blending, altering brightness or contrast, blurring, etc. In the final phase, a new corpus of approximately 8,000 query videos was provided to determine the final ranking.

\subsection{Ablations}
We provide a step-wise comparison with the baseline which only uses ISC-dt1 model and score normalization to extract embedding. In addition, TN is used for localization for Matching Track evaluation. The evaluation was performed on the training dataset, and the \textmu AP for both tracks are shown in Table~\ref{tab:ablation}. Multi-crop and Consistency Weighting improve \textmu AP by a large margin, and Temporal Concat also has some positive effects.
% For the matching track, we further apply a voting ensemble to improve the final result, shown in Table~\ref{tab:ablation-match}.

% \begin{table}[t]
% \centering
% \begin{tabular}{ccccc}
% \toprule
% Multi-crop + Temporal Concat & Ensemble & \textmu AP \\
% \midrule
% &  & 0.5313 \\
% \checkmark &  & 0.8145 \\
% \checkmark & \checkmark & 0.8355 \\
% \bottomrule
% \end{tabular}
% \caption{Ablation from baseline to our final solution for matching track.}
% \label{tab:ablation-match}
% \end{table}

\subsection{VSC2022 Results}
The final VSC2022 results are shown in Table~\ref{tab:lb_descriptor} and Table~\ref{tab:lb_matching}. There are 344 participants in the descriptor track and 212 participants in the matching track. With various tricks that exploit copy video characteristics, our simple pipeline achieved 3rd place for both tracks. 

\section{Conclusion}
This paper presents a simple yet powerful pipeline for video copy detection. We incorporate several tricks, such as TTA, Consistency Weighting, and Temporal Concat. Experiments show such tricks significantly improve detection performance. We achieved 3rd place for both the descriptor track and matching track of the VSC 2022.

% \section*{Acknowledgements}

% \clearpage
{\small
\bibliographystyle{ieee_fullname}
\bibliography{egbib}
}

\end{document}